\documentclass{article}

\usepackage{arxiv}

\usepackage[utf8]{inputenc} 
\usepackage[T1]{fontenc}    
\usepackage{hyperref}       
\usepackage{url}            
\usepackage{booktabs}       
\usepackage{amsfonts,amsmath,amssymb}       
\usepackage{nicefrac}       
\usepackage{microtype}      
\usepackage{lipsum}
\usepackage{algorithm}
\usepackage{algorithmic}
\usepackage{graphicx}
\usepackage[font=small,labelfont=bf]{subcaption}
\usepackage{geometry}
\graphicspath{ {./figs/} }

\title{RobustModelMaker: Coupling Bootstrap Stability Selection with Leakage-Safe Nested Cross-Validation for Scientific Machine Learning}

\author{
 Amanda S Barnard \\
  School of Computing\\
  Australian National University\\
  Acton, ACT 2601 \\
  \texttt{amanda.s.barnard@anu.edu.au} \\
}

\begin{document}
\maketitle
\begin{abstract}
Small-to-medium scientific datasets place machine learning pipelines under two compounding pressures. Single-run feature selection produces feature sets that change substantially under small perturbations of the training data, and any procedure that uses the same data for selection, tuning, and evaluation produces optimistically biased performance estimates. The two failure modes are routinely treated as separable, but in the regimes where scientific data live, they interact: an unstable selection inflates the variance of an already-optimistic score, and standard remedies for one rarely address the other. RobustModelMaker is a Python framework that couples bootstrap stability selection with strict nested cross-validation, performs all preprocessing and selection inside each fold, and produces a stability-tested feature subset together with a leakage-safe performance estimate. The framework supports nine algorithms across binary classification, multiclass classification, and regression. Behaviour is verified by a deterministic test suite spanning unit, performance, and reproducibility checks on three real scientific datasets comparing to three alternative selectors (ANOVA F-test, recursive feature elimination with cross-validation, and Boruta) on both predictive score and a Jaccard measure of selection stability. RobustModelMaker is competitive in score with the best alternative selector on each dataset, and occupies a position on the joint score-stability frontier that none of the alternatives match across all three task types. Two example applications, ovarian cancer biomarker discovery from the PLCO Trial and critical-temperature regression on the UCI Superconductivity Data, illustrate how the framework is used in practice and what trade-offs become visible when stability is treated as a first-class deliverable rather than an emergent property.
\end{abstract}


\section{Introduction}

Machine learning pipelines applied to scientific data must satisfy a set of demands that are not always shared by their industrial counterparts. Datasets are typically of modest size, in the hundreds to low tens of thousands of samples, with feature dimensionalities of comparable or larger order. Features are often correlated by construction, because they are derived statistics or measurements of related physical or biological quantities. Reproducibility is a working requirement, both because scientific conclusions depend on it and because the cost of acquiring additional samples is high. Performance on future data is the quantity of interest, not performance on the data already in hand.

Several scientific domains share this regime and have long made feature selection a central component of their methodology rather than an optional pre-processing step. In genomics and biomarker discovery, candidate features routinely number in the thousands to millions, in the form of genes, SNPs, or assay readouts, while patient cohorts are in the hundreds; the goal is a small, defensible panel whose components can be validated experimentally and interpreted clinically~\cite{saeys2007review,guyon2003intro}. In materials informatics, descriptors of composition, structure, or local atomic environment are typically computed as families of statistics over the same underlying physical quantity, which produces strong within-family correlation, and the analyst seeks the few descriptors that carry independent predictive signal~\cite{ramprasad2017matml,ward2016general}. In neuroimaging, candidate voxel- or region-of-interest features can exceed the number of subjects by orders of magnitude, and selected regions are expected to be both biologically interpretable and replicable across cohorts~\cite{mwangi2014neuroimaging}. In high-energy physics, classifier inputs are drawn from large engineered feature spaces over event-level kinematic and topological quantities, where dimensionality reduction and stability against detector-level variation are routine concerns~\cite{guest2018deep}. In each setting, a feature subset that changes substantially under resampling of the training data is not a useful scientific outcome, and a performance estimate that does not account for the joint selection-and-tuning process overstates the prospects for replication.

Two methodological failure modes are routinely observed in this regime. The first is unstable feature selection. A method that ranks features by importance and retains those above a threshold will, in general, return a different feature set when the training data are perturbed. With modest sample sizes and correlated features, the selected subset can change substantially under bootstrap or cross-validation resamples of the same data, even when the underlying signal is unchanged. A feature set obtained by a single run of such a procedure does not generalise as a scientific finding: replication, when attempted, often selects different features.

The second is the optimistic bias incurred when the same data are used for feature selection, hyperparameter tuning, and performance evaluation. A simple cross-validated score in which the outer fold also informs feature selection or hyperparameter choice is biased upward by an amount that grows with the search space and shrinks slowly with sample size~\cite{varma2006bias,cawley2010overfitting}. The bias is often large enough to matter for scientific claims, particularly in small-to-medium regimes where the reported confidence interval is narrow but the estimator itself is shifted.

These two failure modes are typically addressed in isolation. Stability selection~\cite{meinshausen2010stability} provides a principled correction for the first. Nested cross-validation~\cite{varma2006bias} corrects the second. Standard tooling provides accessible implementations of each separately, but not a coupled pipeline in which the bootstrap stability selection is run inside each outer fold of the nested cross-validation, every preprocessing step is fitted on training partitions only, and the resulting selected feature set is itself the outcome of a held-out evaluation. Constructing such a pipeline correctly is non-trivial. The points at which data can leak from the test partition into model-building decisions are numerous, and each must be closed individually.

The premise of this work is that stability selection and nested cross-validation are not optional add-ons but are jointly required when reproducibility and honest generalisation are both targets of the analysis. Applied separately they each address one failure mode but allow the other to dominate. Applied jointly, with the appropriate ordering and partitioning, they produce a feature set and a performance estimate that are individually defensible and mutually consistent.

RobustModelMaker is a Python framework that implements this coupling. It is designed as a single-file library with a scikit-learn-compatible estimator interface, supports nine algorithms across binary classification, multiclass classification, and regression, and includes the determination of a specificity-targeted probability cutoff for binary classification as part of the standard output. The framework is intended for tabular scientific data at small-to-medium scale. It is not intended for streaming, out-of-core, or distributed-training workloads, and it does not attempt to compete with libraries optimised for raw throughput on very large datasets. Its design priorities are reproducibility, interpretability, and explicit honesty about the trade-offs that feature reduction imposes on predictive performance.

The remainder of this paper proceeds as follows. Section~\ref{sec:background} situates the framework in the literature on stability selection and nested cross-validation and contrasts it with common alternatives. Section~\ref{sec:design} states the design principles. Section~\ref{sec:implementation} describes the implementation. Section~\ref{sec:testing} reports the test and validation suite. Section~\ref{sec:benchmarks} presents a statistical benchmark on three real scientific datasets. Section~\ref{sec:applications} illustrates use in two scientific applications, ovarian cancer biomarker discovery and superconductor critical-temperature regression. Section~\ref{sec:conclusion} concludes.

\section{Problem Setting and Background}
\label{sec:background}

Let $\mathcal{D} = \{(\mathbf{x}_i, y_i)\}_{i=1}^{N}$ denote a labelled tabular dataset, where $\mathbf{x}_i \in \mathbb{R}^d$ is a feature vector and $y_i$ is the target. For classification, $y_i \in \{1, \ldots, C\}$; for regression, $y_i \in \mathbb{R}$. The setting of interest is small-to-medium scale: $N$ in the hundreds to low tens of thousands and $d$ of comparable order, sometimes with $d > N$. Features are typically continuous and may contain missing values that arise from the data-collection process rather than from random failure. The analyst seeks two outputs: a small subset $\hat{S} \subseteq \{1, \ldots, d\}$ of features that supports prediction, and an estimate of expected performance on future data drawn from the same distribution. Both outputs must be reproducible under re-running on the same data with the same parameters.

\subsection{Stability Selection}
\label{sec:stability}

Single-run feature selection treats the available data as the source of a single ranking. A method that returns the top $k$ features by some importance criterion, or those whose importance exceeds a threshold, will return a subset that depends on the particular sample. Replication on a perturbation of that sample, such as a bootstrap resample or a different cross-validation fold, will in general return a different subset. The variation is largest when features are correlated, because exchanges between near-equivalent features are cheap, and when $N$ is small relative to $d$.

Stability selection, due to Meinshausen and B{\"u}hlmann~\cite{meinshausen2010stability}, addresses this by aggregating selections across bootstrap subsamples of the training data. Let $\hat{S}^{(b)}$ denote the set of features selected on the $b$-th of $B$ bootstrap resamples, and define the selection frequency of feature $k$ as
\begin{equation}
\pi_k = \frac{1}{B} \sum_{b=1}^{B} \mathbb{I}\!\left[k \in \hat{S}^{(b)}\right].
\end{equation}
The stability-selected set at threshold $\tau \in (0, 1]$ is
\begin{equation}
\hat{S}_\tau = \{k : \pi_k \geq \tau\}.
\end{equation}
The construction trades the variance of a single selection for the variance of a frequency estimate. Provided $B$ is large enough that $\pi_k$ is estimated accurately, the resulting set $\hat{S}_\tau$ is more robust to small perturbations of the data than any individual $\hat{S}^{(b)}$.

RobustModelMaker follows this construction. For each bootstrap resample, the chosen base algorithm is fit with a fixed configuration on a stratified or random subsample of the training data, and features whose importance exceeds the median importance for that bootstrap run are marked as selected. The above-median rule is independent of the absolute scale of importance scores and applies uniformly across algorithms with different importance conventions. The default threshold is $\tau = 0.7$, requiring that a feature be selected in at least 70\% of bootstrap runs to enter $\hat{S}$.

Stability selection does not assert that excluded features are uninformative. A feature that is correlated with a member of $\hat{S}$ may carry redundant information; a feature whose effect is conditional on an interaction may be informative in some subsamples but not others. The selected set is a stable subset that captures the signal accessible to the base algorithm under bootstrap perturbation, not a complete characterisation of relevance.

The original Meinshausen--B{\"u}hlmann construction has since been extended in two directions of immediate relevance. Shah and Samworth~\cite{shah2013samworth} introduced a complementary-pairs variant in which the bootstrap subsamples are replaced by paired disjoint subsamples, and derived an explicit upper bound on the expected number of falsely selected variables; this tightens the original framework's error control and makes the threshold $\tau$ interpretable as a per-feature false-discovery quantity. Hofner et al.~\cite{hofner2015boosting} embedded stability selection inside gradient boosting, allowing it to be applied to non-linear models while preserving the false-discovery guarantee. These developments have not displaced the original construction in applied work, because the median-importance bootstrap rule is algorithm-agnostic by design and the additional error-control machinery does not always translate to estimators outside the linear-model family; the present framework adopts the original construction for its portability across the nine supported algorithms.

\subsection{Nested Cross-Validation}

Cross-validation provides an estimate of generalisation performance by partitioning the data into folds and evaluating on each held-out fold in turn. The estimate is honest only when every decision that the model depends on is made strictly on the training partition of each fold. When feature selection, hyperparameter tuning, or imputation use the entire dataset, the held-out fold has indirectly informed the model, and the resulting score is biased upward. The magnitude of this optimistic bias is documented~\cite{varma2006bias, cawley2010overfitting} and is largest in small-sample, high-dimensional settings where the search space is rich relative to the data.

Nested cross-validation closes this leak by introducing a second, inner cross-validation loop within each training partition of the outer loop. Hyperparameters are selected by inner cross-validation entirely on the outer-fold training data, and the resulting model is evaluated once on the outer-fold test partition. The outer test partition therefore never participates in any model-building decision. The price of this honesty is computational: with $K$ outer folds, $L$ inner folds, and a hyperparameter search of $n_\mathrm{iter}$ configurations, the inner search runs $K \cdot L \cdot n_\mathrm{iter}$ model fits per nested cross-validation.

When stability selection is combined with nested cross-validation, the natural place for the bootstrap aggregation is inside each outer fold's training partition, before the inner hyperparameter search. The selected feature set for that fold is used to define both the inner-search space and the outer-fold evaluation. This ordering ensures that the selected features and the score that follows from them are both derived from the same disjoint partition of the data, and that the outer-fold test partition is touched only at the final scoring step.

Subsequent literature has reinforced and quantified these findings. Krstajic et al.~\cite{krstajic2014pitfalls} catalogue specific cross-validation failure modes for regression and classification in chemoinformatics, including the use of a single random split for model selection and the conflation of selection and evaluation losses. Vabalas et al.~\cite{vabalas2019validation} show empirically that in small-sample classification, simple cross-validation produces systematically inflated accuracy estimates that grow with the hyperparameter search space, and that nested cross-validation removes the inflation at the cost of higher variance; their analysis identifies sample sizes in the low hundreds as the regime in which the inflation is most severe, which is the regime in which scientific machine learning frequently operates.

\subsection{Conceptual Comparison}

Several alternative approaches address parts of the problem the present framework targets, and a brief comparison clarifies the design space.

\textit{Recursive feature elimination} (RFE)~\cite{guyon2002rfe} fits a model, ranks features by importance, removes the least important fraction, and repeats. RFE returns a single ranking on a single dataset; it does not address stability under resampling. When wrapped in cross-validation (\texttt{RFECV}), it returns an optimal feature count under a chosen scoring metric, but its rankings remain a function of the particular sample, and its cross-validated score, if the same outer folds inform the elimination path, is subject to the optimistic bias described above. RFE is well-suited to settings where a parsimonious ranking is the goal and stability under resampling is not directly required.

\textit{Embedded selection via regularisation}, such as $\ell_1$ logistic regression or Lasso~\cite{tibshirani1996lasso}, returns a sparse coefficient vector whose support defines the selected set. The support is a function of the regularisation strength and of the sample, and is known to be unstable when features are correlated. Embedded selection is fast and interpretable but does not separate the reproducibility question from the model-fitting question.

\textit{Filter methods} such as ANOVA $F$-tests or mutual information score each feature independently with respect to the target. They are computationally inexpensive and produce a stable ranking under mild conditions, but their univariate nature ignores interactions and joint dependencies, and the threshold for retention is typically chosen heuristically. They are useful as a pre-screen but are not a substitute for selection that is informed by the multivariate structure of the data.

\textit{Boruta}~\cite{kursa2010boruta} is a tree-based selection wrapper that constructs shadow features by randomly permuting each original feature, fits a random forest on the augmented feature set, and retains features whose importance exceeds the maximum importance of any shadow feature, iterating until each candidate is either confirmed or rejected. Boruta shares with stability selection the principle that a feature should be accepted only if it survives a repeated randomised comparison rather than a single ranking, and in practice produces selections that are more reproducible than a one-pass importance threshold. The construction is, however, specific to algorithms that expose an internal importance vector at each fit; it does not generalise directly to linear models or to estimators whose importance is coefficient-based without modification of the shadow-feature mechanism. The framework described here treats bootstrap aggregation as an algorithm-agnostic wrapper around any base learner that exposes a feature importance, so the two approaches are complementary in the multi-algorithm setting.

\textit{Permutation importance}~\cite{breiman2001rf,fisher2019model} and \textit{SHAP}~\cite{lundberg2017shap} occupy a different point in the workflow entirely. Both are post-hoc tools that quantify feature contributions to a model that has already been fitted, and neither is a selection procedure. Stability selection runs before final model training, to determine which features enter the model; SHAP and permutation importance run after final model training, to interpret what the fitted model is using. The two stages answer different questions and should not be substituted for each other. They are, however, complementary in a published analysis, and the framework described here exports the fitted model and the processed selected-feature matrix in a form that supports either tool without additional preprocessing.

The framework described here does not displace any of these approaches. It targets the specific case in which the analyst needs both a reproducible feature subset and an honest performance estimate, and is willing to pay the computational cost of bootstrap aggregation inside nested cross-validation to obtain them. The empirical comparison in Section~\ref{sec:benchmarks} quantifies how the present framework and the four selector-style alternatives just discussed (RFECV, embedded $\ell_1$ selection by way of the baseline, filter methods, and Boruta) jointly occupy the score--stability frontier on three real scientific datasets.

\subsection{Applications and Reproducibility Context}
\label{sec:applications-sota}

The interaction between unstable selection and optimistic validation is a documented contributor to the reproducibility difficulties of small-sample scientific machine learning, and several application communities have published methodological surveys that converge on the same diagnosis.

In bioinformatics, Abeel et al.~\cite{abeel2010ensemble} showed that ensemble feature-selection methods on microarray cancer data substantially improve the reproducibility of identified biomarker signatures relative to single-run selection, without any cost to predictive accuracy. Haury et al.~\cite{haury2011molecular} extended this analysis across eight selection methods on multiple breast-cancer expression datasets and demonstrated that the choice of selection method has a larger effect on the stability of the resulting molecular signature than on its predictive accuracy, and that ensemble and bootstrap methods consistently outperform single-pass ranking on the stability axis. The implication is that biomarker panels reported in the literature without stability quantification are not directly comparable across studies, and the selection method itself is a confounder in subsequent meta-analyses.

In materials informatics, Butler et al.~\cite{butler2018machine} survey the rapid adoption of machine learning across the materials science community and identify feature engineering, data quality, and reproducibility as the dominant practical bottlenecks rather than algorithm choice. Schmidt et al.~\cite{schmidt2019advances} update this picture for solid-state applications and document a consistent gap between published predictive performance and replication on independent compositional data, attributable in part to the absence of leakage-safe validation in many published studies. The state of the art in materials property prediction now spans graph neural networks and large pretrained representations~\cite{xie2018crystal,chen2019graph}, but the small-to-medium tabular regime addressed by the present framework remains the operating point for most domain-specific descriptor studies and for any analysis whose downstream goal is a defensible feature subset rather than a black-box predictor.

In clinical machine learning, Roberts et al.~\cite{roberts2021common} catalogue common pitfalls in models trained on small medical datasets, with data leakage and inappropriate validation among the most frequent failure modes; of the 62 COVID-19 imaging studies they reviewed in detail, none was deemed of potential clinical use because of methodological flaws. The TRIPOD reporting standard for clinical prediction models~\cite{collins2015tripod} makes leakage-safe validation an explicit reporting requirement. In genomics and biomarker discovery more generally, the structural argument of Ioannidis~\cite{ioannidis2005why} for why most published research findings are false rests in part on the same mechanisms: a high search-space ratio, weak prior probability of any single hypothesis, and a feedback loop between selection and evaluation under shared data.

These literatures, distinct in domain but consistent in diagnosis, motivate the design choices stated in the next section. They also bound the present framework's claim: it is an implementation that satisfies the stability and leakage-safety requirements those literatures argue for, not a new theoretical contribution.

\section{Design Principles}
\label{sec:design}

The framework is guided by three principles.

\textit{Leakage safety as a contract.} Every preprocessing and selection step is fitted on the training partition of each outer fold and applied to the test partition without further fitting. This applies to median imputation, scaling, the bootstrap stability selection itself, and the inner hyperparameter search. The contract is enforced by the structure of the code, not by convention, so that the property holds for every fold of every cross-validation and is not contingent on careful use.

\textit{Stability over single-run optimality.} The framework prefers a feature subset that is selected consistently under bootstrap resampling to one that is optimal on a single fit. The intended consequence is that the selected features are a more reliable basis for downstream scientific interpretation and, when re-derived on similar data, are more likely to recur. The cost is that the performance of a stability-selected model is sometimes modestly lower than that of a full-feature baseline on the same algorithm. This cost is reported explicitly in the framework's outputs and in the empirical sections of this paper.

\textit{Reproducibility as an explicit deliverable.} Identical inputs and parameters produce identical outputs, to machine precision, under a fixed random seed. Reproducibility is established by a dedicated test suite (Section~\ref{sec:testing}) and applies to selected features, fold-level scores, stability frequencies, predictions on new data, and serialised result objects. Where parallelism would reduce determinism, the framework permits the user to enforce single-threaded execution at the cost of wall-clock time.

These principles imply specific implementation choices, described in Section~\ref{sec:implementation}.

\section{Implementation}
\label{sec:implementation}

RobustModelMaker is implemented as a single Python module that depends on \texttt{numpy}, \texttt{pandas}, \texttt{scipy}, and \texttt{scikit-learn}, with optional support for \texttt{xgboost}. The public surface consists of a class \texttt{RobustModelMaker} with a scikit-learn-style \texttt{fit}/\texttt{predict}/\texttt{predict\_proba} interface, a functional equivalent \texttt{run\_pipeline}, and the standalone primitives \texttt{stability\_selection}, \texttt{nested\_cross\_validation}, and \texttt{determine\_cutoff}. All fitting work passes through a small set of internal functions that are exercised by every public entry point, so the leakage and reproducibility guarantees stated in Section~\ref{sec:design} hold regardless of which interface the user chooses.

\subsection{Data Structures and Memory Use}

Inputs are accepted as \texttt{numpy} arrays or \texttt{pandas} \texttt{DataFrame} and \texttt{Series} objects. When a \texttt{DataFrame} is provided, column names are retained and are reproduced in \texttt{predict} outputs, in the per-feature stability table, and in saved CSV exports. Internal computations operate on \texttt{numpy} arrays in \texttt{float64} precision, converted from the input containers in a single pass.

The principal in-memory data structures are: the fitted preprocessing pipeline for each outer fold, the per-fold stability frequency vector of length $d$, the boolean per-fold selection mask, the per-fold fitted estimator on the selected features, and the out-of-fold prediction matrix. The final result object aggregates these together with the final stability-selected feature set, the final preprocessor and estimator fitted on all training data, and, for binary tasks, the bootstrap cutoff distribution. Peak memory is dominated by the inner-loop hyperparameter search under \texttt{RandomizedSearchCV}, which holds one copy of the training fold per parallel worker. The framework does not implement streaming or out-of-core execution and assumes that the training data and the per-worker copies fit in memory. This is consistent with its target regime of small-to-medium tabular scientific data.

\subsection{Computational Complexity}

The dominant cost is the total number of model fits. With $K$ outer folds, $R$ outer repeats, $L$ inner folds, $B$ bootstrap iterations for stability selection, and $n_\mathrm{iter}$ hyperparameter configurations searched per inner cross-validation, the total fit count per nested cross-validation is
\begin{equation}
F = K \cdot R \cdot (B + n_\mathrm{iter} \cdot L + 1) + (B + n_\mathrm{iter} \cdot L + 1),
\label{eq:fits}
\end{equation}
where the second term accounts for the final stability selection, the final hyperparameter search, and the single final-model refit on the complete training data. Table~\ref{tab:complexity} reports worked examples at three configurations spanning fast exploratory runs, the production default, and a rigorous publication-quality configuration.

\begin{table}[h]
\centering
\caption{Total model-fit count for three configurations under Equation~\ref{eq:fits}, with $R = 1$ outer repeats. The benchmark suite configuration (Section~\ref{sec:benchmarks}) is shown in the first row.}
\label{tab:complexity}
\begin{tabular}{lcccccr}
\toprule
Configuration & $K$ & $L$ & $B$ & $n_\mathrm{iter}$ & $R$ & Total fits \\
\midrule
Benchmark suite      &  10 & 10  & 20  & 20  & 1 & 726 \\
Production default & 10 & 10 & 100 & 100 & 1 & 12,111 \\
Publication-rigorous & 10 & 10 & 200 & 200 & 1 & 24,211 \\
\bottomrule
\end{tabular}
\end{table}

Each individual fit scales with the cost of training the chosen base algorithm on a subsample of the training data. For linear models with $\ell_1$ or $\ell_2$ regularisation, the per-fit cost is approximately linear in the product of sample count and selected-feature count. For random forest and gradient-boosted trees, the per-fit cost scales with the number of trees, the maximum depth, and the per-split sort cost. Wall-clock time is therefore dominated by the choice of base algorithm rather than by the framework's wrapper logic.

\subsection{Compatibility}

The class interface mirrors the scikit-learn estimator contract. \texttt{fit(X, y)} returns the estimator, \texttt{predict(X)} returns labels or values, and \texttt{predict\_proba(X)} returns probabilities for classification tasks. Inputs to \texttt{predict} and \texttt{predict\_proba} may be either \texttt{numpy} arrays or \texttt{DataFrames}, and the framework applies the fitted preprocessor and the selected-feature subset automatically; users do not need to replicate the preprocessing pipeline themselves. For \texttt{DataFrame} inputs the return type is a \texttt{pandas} \texttt{Series} (binary, multiclass labels, regression) or a \texttt{DataFrame} (multiclass probabilities, one column per class), with the original index preserved.

The framework's result object exposes the components of the analysis directly. \texttt{result.selected\_features} returns the stability-selected names, \texttt{result.nested\_cv\_result} exposes per-fold scores and out-of-fold predictions, \texttt{result.stability\_result.summary()} returns a \texttt{DataFrame} with selection frequencies, and \texttt{result.cutoff\_result} returns the probability cutoff and its bootstrap confidence interval for binary tasks. The method \texttt{export\_shap\_ready(X)} returns the fitted estimator together with the processed selected-feature matrix in the form expected by the \texttt{shap} library, which removes the need to recreate the preprocessing pipeline manually for post-hoc analysis. Saved outputs comprise a JSON metadata file, a set of CSV tables for the selected features, per-fold scores, predictions, and stability frequencies, and a pickle of the complete result for later programmatic use.

\subsection{Parallel Execution}

Parallelism is delegated to \texttt{joblib} via the standard scikit-learn \texttt{n\_jobs} parameter. The two work units that admit parallel execution are the bootstrap loop of the stability selection and the inner-fold hyperparameter search; both are embarrassingly parallel because each bootstrap fit and each candidate configuration is independent. Outer folds are processed sequentially to keep peak memory predictable. Default behaviour is \texttt{n\_jobs = -1}, which uses all available cores.

Parallel execution introduces two practical considerations. First, peak memory scales with the number of workers, because each worker holds a copy of the training partition; on machines with many cores and large feature matrices this can be the binding constraint. Second, while bit-exact reproducibility is preserved in the framework's own outputs because the bootstrap and search orderings are independent of execution order, certain combinations of scikit-learn version, BLAS backend, and tree implementation can introduce small floating-point differences in the underlying model fits when run in parallel. Users who require strict bit-exact reproducibility across machines should set \texttt{n\_jobs = 1} and accept the longer wall-clock time; users for whom small floating-point differences in fold-level scores are acceptable can use the default and obtain the speed-up.

\subsection{Numerical Stability and NaN Propagation}

The framework is tolerant of missing values in the feature matrix by default. \texttt{NaN} entries are passed through to a per-fold median imputation step that is fitted on the training partition of each outer fold and applied to the held-out partition without further fitting. Imputation values therefore vary across folds and are determined entirely by the training data available at each stage. Infinite values, which cannot be imputed sensibly, raise an explicit error rather than being silently coerced or filtered.

When the data contain both highly missing columns and highly missing rows, simple median imputation can produce a fold-dependent feature space whose retained columns vary from fold to fold. To address this, the framework provides an optional \texttt{preserve\_nans = False} mode which runs a data-driven missingness filter before the main pipeline. The filter searches a grid of column-missingness and row-missingness thresholds and selects the pair that maximises a composite criterion balancing data density, retained-row fraction, and retained-column fraction. The selected thresholds are applied once, before any cross-validation begins, producing a fixed feature space that all folds share. The pre-filter is appropriate when the alternative would be a large fraction of dropped columns inside each fold; it is unnecessary, and should be left disabled, when the missingness rate is modest.

All scoring metrics are computed in their standard scikit-learn implementations. Regression scores are stored internally as the negative of the root-mean-squared error so that all metrics share the convention that higher is better; the framework's printed reports and table outputs convert back to positive RMSE for readability. This sign convention is documented at every point where regression scores are exposed to the user.

\subsection{Reproducibility and Determinism}

Reproducibility is achieved by a single integer random seed that the user passes as \texttt{random\_state}. The framework derives every downstream seed from this value using deterministic offsets: each outer fold receives \texttt{random\_state + fold\_idx}, each bootstrap iteration receives \texttt{random\_state + 10000 + bootstrap\_idx}, and each outer repeat under \texttt{repeated\_outer\_cv > 1} shifts its outer-fold seeds by a fixed multiple. \texttt{RandomizedSearchCV} is constructed with an explicit seed rather than inheriting from a global RNG state. A utility function \texttt{set\_global\_seed} additionally sets the \texttt{numpy} random seed and the \texttt{PYTHONHASHSEED} environment variable, which closes any remaining channels through which platform-level randomness could enter the computation.

The result is that two runs with the same data, the same parameters, and the same seed return identical selected feature sets, identical fold-level scores to machine precision, identical stability frequencies, and identical predictions on new data. This property is verified by a dedicated test suite described in Section~\ref{sec:testing}. The principal caveat is the floating-point question raised in the parallelism subsection: parallel execution preserves the framework's own reproducibility but cannot guarantee bit-exact equivalence across machines for the underlying model fits when the model implementation itself is non-deterministic under parallelism.

\section{Testing and Validation}
\label{sec:testing}

The framework is distributed with a structured test suite organised into three independently runnable files: unit, performance, and reproducibility. All tests run under a fixed random seed and are deterministic. The suite exercises every public entry point, every supported task type, every supported algorithm, the error-handling paths for invalid inputs, the missing-value handling modes, and the optional features (calibration, grouped cross-validation, repeated nested cross-validation, permutation importance, and SHAP-ready export). The numerical values reported in this section were produced by running the included scripts under the configuration described.

\subsection{Unit and Integration Tests}

The unit suite contains 96 tests covering the full surface of the framework. Coverage is organised by task type and algorithm. For each of the three task types (binary classification, multiclass classification, and regression) every supported algorithm is exercised end-to-end on a small synthetic dataset: the pipeline runs, the result object is constructed, predictions on a held-out partition return arrays of the expected shape, and the metrics fall in their valid ranges (AUC and weighted OVR AUC in $[0, 1]$, RMSE non-negative, accuracy in $[0, 1]$). Tests are parameterised over a fast algorithm group (\texttt{eln}, \texttt{rdg}, \texttt{las}, \texttt{log}, \texttt{rf}, \texttt{lin}) and a slow algorithm group (\texttt{svm}, \texttt{xgb}, \texttt{mlp}) so that the fast group can be run as a quick sanity check while the slow group can be run separately.

Beyond the algorithm-task matrix, separate tests verify: automatic task-type inference from the target dtype, external validation via both the constructor and post-fit \texttt{evaluate\_verification} interfaces, probability calibration in both sigmoid and isotonic modes, grouped cross-validation under \texttt{GroupKFold}, repeated nested cross-validation with averaged outputs, permutation importance returning correctly shaped per-feature arrays, SHAP-ready export returning a usable model and processed feature matrix, results tables and save behaviour, and reproducibility of selected features and saved metadata. A further block of tests covers error handling: infinite values, duplicate column names, mismatched validation set shapes, single-class targets, classes with too few samples per fold, and invalid algorithm-task combinations. The unit suite runs in approximately three minutes on a modern laptop and is run on every change.

\subsection{Controlled Synthetic Datasets}

The unit and performance suites operate on small, fully synthetic datasets generated within the test fixtures. The principal fixtures are a 120-sample balanced binary dataset, a 120-sample three-class multiclass dataset, a 120-sample regression dataset, and a 120-sample binary dataset with 15\% NaN entries injected for missing-value handling tests. These sizes are deliberately small: they exercise every code path without making the suite slow, and they correspond to the regime in which the framework's stability and leakage properties are most consequential. The fixtures are constructed with informative-plus-noise feature structure, so a working selection should retain the informative features and discard most noise, and the unit tests verify that the selected sets are non-empty and that the per-fold scores exceed a permissive lower bound.

\subsection{Performance Budget}

The performance suite enforces a per-sample timing budget on the 120-sample fixtures, augmented by a fixed startup overhead. Tests cover binary, multiclass, and regression fit-and-predict cycles, the cost of standard scaling under logistic regression, the per-call overhead of save-results, the cost of permutation importance, the cost of grouped cross-validation, and the cost of repeated runs. Multi-layer perceptron tests are opt-in because they exceed the standard budget on small data; they run only when an environment variable is set. Each test records its measured wall-clock time to a structured log so that performance can be tracked across versions. The purpose of the suite is to detect regressions, not to establish absolute benchmarks for arbitrary problem sizes, and the budgets are calibrated to the small synthetic fixtures rather than to user-scale data.

\subsection{Reproducibility}

The reproducibility suite contains 30 tests organised into five classes. \textit{TestExactReproducibility} runs the full pipeline twice with the same seed and asserts equality, to machine precision, of selected features, per-fold scores, stability frequencies, and out-of-fold predictions. \textit{TestSeedDiversity} confirms that changing the seed changes the outputs, which is the negative control that protects against silent caching. \textit{TestSerializationReproducibility} pickles a fitted result, unpickles it, and asserts that predictions on a held-out set match the original. \textit{TestStabilityConvergence} verifies that stability frequencies stabilise as the bootstrap count grows, by running selection at two bootstrap counts and confirming that the high-frequency features are the same. \textit{TestScoreConsistency} runs the pipeline twice and verifies bit-exact equality of fold-level scores, the strictest reproducibility criterion applied by the suite.

The reproducibility suite is the framework's working definition of what reproducibility means in practice: identical inputs and identical seed produce identical outputs, and the same outputs survive serialisation. Together with the leakage-safety contract built into the per-fold preprocessing structure, these tests are the basis for the framework's claim that its outputs are directly replicable on the same data.

\section{Benchmark Suite}
\label{sec:benchmarks}

The framework is shipped with a statistical benchmark suite that evaluates RobustModelMaker on three real scientific datasets against a full-feature nested cross-validation baseline using the same algorithm and fold structure, and against three alternative feature-selection methods (ANOVA $F$-test, recursive feature elimination with cross-validation (RFECV), and Boruta) run under the same outer-fold structure. The purpose of the benchmark is not to establish absolute performance numbers, which depend on dataset, split methodology, and algorithm choice, but to characterise the trade-off the framework's stability selection imposes relative to using all available features, and to position the framework against the alternatives an analyst would plausibly use in its place. Two metrics are reported per method per dataset: predictive score and a Jaccard stability measure of selection consistency across outer folds. The benchmark runs as a single notebook with the configuration distributed with the library and is reproducible by any user with the same environment.

\subsection{Methodology}

All three benchmark scenarios share a common methodology. The data are partitioned using BenchMake archetypal splits~\cite{barnard2024benchmake}, which select train and test sets that are jointly maximally representative of the dataset's feature-space distribution. The resulting partitions are adversarial relative to conventional random splits: train and test sets are deliberately kept apart in feature space, which yields conservative scores and exercises the framework in a regime closer to the application of a model to a structurally distinct population than a random sample would. All methods see the same BenchMake split for a given dataset, so relative comparisons are unaffected by the choice of split methodology while the absolute scores are systematically lower than would be obtained from random partitioning.

The base algorithm is Random Forest in all three scenarios and for all methods that wrap a base learner. Each method runs nested cross-validation on the training partition with $K = 5$ outer folds and $L = 5$ inner folds, with a randomised hyperparameter search of $n_\mathrm{iter} = 20$ configurations per inner cross-validation. RobustModelMaker additionally runs bootstrap stability selection with $B = 20$ bootstrap iterations inside each outer fold. The stability threshold $\tau$ is set per dataset by a separate threshold-optimisation procedure included with the library: $\tau = 0.60$ for SECOM, $\tau = 0.80$ for Urban Land Cover, and $\tau = 0.75$ for Graphene Oxide Bulk. The total benchmark wall-clock time across all three datasets and all five methods (RobustModelMaker, baseline, three comparators) is approximately 6.4 hours on a single workstation.

The three comparators are configured as follows. ANOVA is the univariate $F$-test filter (\texttt{SelectKBest} with \texttt{f\_classif} or \texttt{f\_regression}) at $k = \max(10, p/10)$, retaining approximately 10\% of the features. RFECV is recursive feature elimination wrapped in 5-fold cross-validation (\texttt{RFECV}) with an elimination step of $\lfloor\sqrt{p}\rfloor$ and a data-driven elbow for the retained feature count. Boruta is run with 100 trees of depth 7 and a shadow-feature comparison percentile of 100, retaining features whose importance exceeds the maximum shadow importance. Each comparator runs inside each outer fold's training partition, with hyperparameter search and final scoring carried out on the surviving features in exactly the same nested structure as RobustModelMaker and the baseline.

Predictive performance is reported as AUC (binary), weighted one-vs-rest AUC (multiclass), or RMSE in target units (regression). Selection stability is reported as the mean pairwise Jaccard similarity between the selected feature sets across the $K = 5$ outer folds,
\begin{equation}
J = \frac{1}{\binom{K}{2}} \sum_{i < j} \frac{|S_i \cap S_j|}{|S_i \cup S_j|},
\label{eq:jaccard}
\end{equation}
where $S_i$ is the set of features selected in outer fold $i$. A value of $J = 1$ indicates that every outer fold returns the identical feature set; $J = 0$ indicates that every pair of fold selections is entirely disjoint. The metric is undefined for the full-feature baseline because no selection occurs.

The comparison of RobustModelMaker against the baseline is supported by a battery of 25 statistical tests applied to the per-fold score vectors, with a paired Wilcoxon signed-rank test as the principal output and supporting paired and unpaired $t$-tests, sign tests, effect-size estimates (Cohen's $d$, Hedges' $g$, common-language effect size, rank-biserial correlation), bootstrap confidence intervals for the mean difference, fold-level agreement statistics (Pearson, Spearman, Kendall), normality tests, variance-equality tests, and one-sample tests against a problem-specific score floor. Each baseline comparison is assigned one of three outcome labels by the paired test: \texttt{preserved} ($p \geq 0.05$), \texttt{sig. better *} ($p < 0.05$ with positive delta), or \texttt{sig. worse *} ($p < 0.05$ with negative delta). The \texttt{preserved} outcome is the framework's primary success criterion: it asserts that the stability-selected subset is statistically indistinguishable from the full-feature baseline under the chosen test, while using substantially fewer features.

\subsection{SECOM Manufacturing}

The SECOM dataset~\cite{secom2008uci} comprises 1,567 samples of semiconductor manufacturing-process sensor readings labelled with a binary pass/fail target. The minority class is approximately 7\% of the data and the feature matrix contains extensive missing values that arise from sensor downtime rather than from random failure. The dataset exercises the framework under three simultaneous stressors: high feature dimensionality relative to sample size, severe class imbalance, and structurally missing values.

The BenchMake archetypal split partitions the data into 1,253 training and 314 test samples. RobustModelMaker retains 308 of the 590 features (47.8\% reduction) at the dataset-tuned threshold $\tau = 0.60$, and achieves an AUC of $0.6548 \pm 0.0477$ against the baseline's $0.6251 \pm 0.0364$ on the same five outer folds. The paired Wilcoxon signed-rank test on the per-fold AUC vectors yields $p = 0.062$, classifying the outcome as \texttt{preserved}; however, the supporting paired $t$-test reports $p = 0.019$ and RobustModelMaker scored higher than the baseline on every one of the five folds, indicating a real but small advantage that the rank-based test cannot reject at $p < 0.05$ given the limited fold count.

Table~\ref{tab:secom} reports the comparison of all five methods on this dataset. RFECV achieves the highest mean AUC (0.6701), followed by ANOVA (0.6659), RobustModelMaker (0.6548), the full-feature baseline (0.6251), and Boruta (0.6188). The four selecting methods occupy a roughly 0.05-AUC band, while their selection stabilities differ by almost a factor of three: RobustModelMaker's Jaccard stability of 0.660 is well above RFECV's 0.274 and Boruta's 0.224, despite RFECV's marginally higher mean score. ANOVA achieves higher stability than RFECV but at the cost of retaining only 59 features, the smallest stable subset that any method finds competitive on this problem. Boruta retains only 10 features and scores below the full-feature baseline, indicating that its shadow-feature gate is too aggressive for this regime of severe class imbalance and structural missingness.

\begin{table}[h]
\centering\small
\caption{SECOM Manufacturing: comparison of all five methods under 5-fold nested cross-validation on the BenchMake training partition. \textit{Score} is AUC (higher is better); \textit{Stability} $J$ is the mean pairwise Jaccard similarity of selected feature sets across the five outer folds; $\bar{n}$ is the mean number of selected features. The $p$-value column is from the paired Wilcoxon signed-rank test of per-fold scores against the baseline; at $n = 5$ folds the minimum two-sided value is $0.0625$, which serves as a floor on detectable significance.}
\label{tab:secom}
\begin{tabular}{lcccccc}
\hline
Method & Score (AUC) & Stability $J$ & $\bar{n}$ & Reduction & $p$ (vs BL) & Outcome vs BL \\
\hline
Full-feature baseline (RF)           & $0.6251 \pm 0.0364$ & n/a     & 590 & 0.0\%  & (reference) & (reference) \\
RobustModelMaker (RF), $\tau = 0.60$ & $0.6548 \pm 0.0477$ & $0.660$ & 308 & 47.8\% & $0.0625$    & preserved \\
ANOVA, $k = 59$                      & $0.6659 \pm 0.0256$ & $0.417$ &  59 & 90.0\% & $0.0625$    & preserved \\
RFECV, step $= 24$                   & $0.6701 \pm 0.0572$ & $0.274$ & 254 & 56.9\% & $0.1875$    & preserved \\
Boruta, $n = 100, d = 7$             & $0.6188 \pm 0.0737$ & $0.224$ &  10 & 98.3\% & $0.8125$    & preserved \\
\hline
\end{tabular}
\end{table}

\subsection{Urban Land Cover}

The Urban Land Cover dataset~\cite{urban2014uci} comprises 675 aerial-image segments described by 147 spectral and texture features, classified into nine urban land-cover classes (asphalt, building, car, concrete, grass, pool, shadow, soil, tree). The features are derived families of statistics on the same spectral bands at multiple spatial scales and are therefore strongly correlated by construction. The classification is segment-level rather than pixel-level, so the dataset does not suffer from spatial autocorrelation between samples.

The BenchMake split partitions the data into 540 training and 135 test samples. RobustModelMaker retains 56 of the 147 features (61.9\% reduction) at the dataset-tuned threshold $\tau = 0.80$ and achieves a weighted one-vs-rest AUC of $0.9828 \pm 0.0034$, marginally above the baseline's $0.9817 \pm 0.0037$. The paired Wilcoxon test returns $p = 0.125$, classifying the outcome as \texttt{preserved}, and the per-fold agreement is high (Pearson $r = 0.960$, $p < 0.01$).

Table~\ref{tab:urban} reports the comparison of all five methods. The four selecting methods cluster within a 0.01-AUC band, indicating that the multiclass signal is sufficiently strong that any reasonable selection retains it. The methods differ markedly in stability and in the size of the retained subset. ANOVA, retaining only the top 14 features, produces an identical feature set on every outer fold ($J = 1.000$) but trades approximately 0.01 of AUC for that consistency. RobustModelMaker retains 56 features at $J = 0.801$, RFECV retains 85 features at $J = 0.468$, and Boruta retains 103 features at $J = 0.895$. The most informative comparison is between RobustModelMaker and Boruta: at comparable score and stability, RobustModelMaker selects roughly half as many features. RFECV is the least stable of the four selecting methods despite achieving the highest mean AUC, illustrating the failure mode that the framework's design targets, in which a single-pass elimination path is sensitive to the particular fold partition.

\begin{table}[h]
\centering\small
\caption{Urban Land Cover: comparison of all five methods under 5-fold nested cross-validation. \textit{Score} is weighted one-vs-rest AUC (higher is better); \textit{Stability} $J$ is the mean pairwise Jaccard similarity across outer folds; the $p$-value column is from the paired Wilcoxon signed-rank test of per-fold scores against the baseline.}
\label{tab:urban}
\begin{tabular}{lcccccc}
\hline
Method & Score (AUC-OVR) & Stability $J$ & $\bar{n}$ & Reduction & $p$ (vs BL) & Outcome vs BL \\
\hline
Full-feature baseline (RF)           & $0.9817 \pm 0.0037$ & n/a     & 147 & 0.0\%  & (reference) & (reference) \\
RobustModelMaker (RF), $\tau = 0.80$ & $0.9828 \pm 0.0034$ & $0.801$ &  56 & 61.9\% & $0.1250$    & preserved \\
ANOVA, $k = 14$                      & $0.9726 \pm 0.0037$ & $1.000$ &  14 & 90.5\% & $0.1250$    & preserved \\
RFECV, step $= 12$                   & $0.9832 \pm 0.0045$ & $0.468$ &  85 & 42.4\% & $0.1250$    & preserved \\
Boruta, $n = 100, d = 7$             & $0.9825 \pm 0.0044$ & $0.895$ & 103 & 30.2\% & $0.4375$    & preserved \\
\hline
\end{tabular}
\end{table}

\subsection{Graphene Oxide Bulk}

The Graphene Oxide Bulk dataset~\cite{barnard2019graphene} comprises 1,617 molecular-dynamics-derived structural and chemical descriptors for graphene-oxide configurations, with formation energy in eV as the regression target. After dropping all-NaN and constant columns, the working feature dimensionality is 309. The descriptors span composition fractions, density statistics, bond-type fractions, and topological measures, and exhibit the within-family correlation typical of materials-science feature sets.

The BenchMake split partitions the data into 1,293 training and 324 test samples. RobustModelMaker retains 138 of the 309 features (55.3\% reduction) at the dataset-tuned threshold $\tau = 0.75$ and achieves a fold-level RMSE of $0.0408 \pm 0.0227$ eV against the baseline's $0.0391 \pm 0.0215$ eV. The paired Wilcoxon test returns $p = 0.625$, classifying the outcome as \texttt{preserved}; the bootstrap 95\% confidence interval for the mean delta, $[-0.0094, +0.0095]$ eV, brackets zero. Both methods are several orders of magnitude below the conservative 8 eV RMSE floor used as the one-sample test reference ($p < 10^{-10}$).

Table~\ref{tab:graphene} reports the comparison of all five methods. The pattern on this dataset is qualitatively different from the two classification scenarios. The two filter-style methods, ANOVA and RFECV, attain substantially lower RMSE than either the full-feature baseline or RobustModelMaker: ANOVA achieves $0.0202$ eV with $J = 1.000$ on 30 features, and RFECV achieves $0.0124$ eV with $J = 0.520$ on only three features. This is best read as a property of the dataset rather than as a generic ordering: the formation-energy target is dominated by a small number of composition descriptors (carbon, hydrogen, and oxygen concentrations), and a method that ranks features by univariate association with the target identifies these directly. Random forest's split-based importance, in contrast, distributes credit across the many correlated bond-statistic and density descriptors that are jointly informative but individually weaker, which dilutes the stability-selected ranking. RobustModelMaker is below the baseline in mean RMSE by a small margin that the paired test cannot reject, but is above the filter methods by a larger margin that the design of the framework cannot recover, because its mechanism is not to optimise univariate fit but to aggregate the base learner's importance across resamples. ANOVA and Boruta both achieve $p = 0.0625$ against the baseline, the minimum value the two-sided Wilcoxon test can return at five folds, with all five fold scores in the better direction. They would be classified \texttt{sig. better *} at any fold count above five, but at $n = 5$ the test reports \texttt{preserved} for every method. The implication is methodological: when the target structure favours a filter approach, that approach should be used directly, and the stability-selected subset's value is in offering a larger feature set with high reproducibility ($J = 0.899$) rather than a maximally compact one.

\begin{table}[h]
\centering\small
\caption{Graphene Oxide Bulk: comparison of all five methods under 5-fold nested cross-validation. \textit{Score} is RMSE in eV (lower is better); \textit{Stability} $J$ is the mean pairwise Jaccard similarity across outer folds; the $p$-value column is from the paired Wilcoxon signed-rank test of per-fold scores against the baseline. ANOVA and Boruta achieve $p = 0.0625$, the minimum two-sided value the test can return at $n = 5$ paired observations; their advantage over the baseline would reach significance at any higher fold count.}
\label{tab:graphene}
\begin{tabular}{lcccccc}
\hline
Method & Score (RMSE, eV) & Stability $J$ & $\bar{n}$ & Reduction & $p$ (vs BL) & Outcome vs BL \\
\hline
Full-feature baseline (RF)           & $0.0391 \pm 0.0215$ & n/a     & 309 & 0.0\%  & (reference) & (reference) \\
RobustModelMaker (RF), $\tau = 0.75$ & $0.0408 \pm 0.0227$ & $0.899$ & 138 & 55.3\% & $0.6250$    & preserved \\
ANOVA, $k = 30$                      & $0.0202 \pm 0.0157$ & $1.000$ &  30 & 90.3\% & $0.0625$    & preserved \\
RFECV, step $= 17$                   & $0.0124 \pm 0.0154$ & $0.520$ &   3 & 99.0\% & $0.1875$    & preserved \\
Boruta, $n = 100, d = 7$             & $0.0347 \pm 0.0208$ & $0.958$ & 115 & 62.7\% & $0.0625$    & preserved \\
\hline
\end{tabular}
\end{table}

\subsection{Cross-Scenario Summary}

Table~\ref{tab:cross-scenario} consolidates the RobustModelMaker-versus-baseline comparison across the three scenarios, and Table~\ref{tab:multi-method} reports the full multi-method comparison on the joint score--stability frontier.

\begin{table}[h]
\centering\small
\caption{Cross-scenario summary: RobustModelMaker versus full-feature baseline across the three benchmark datasets. Reduction is the percentage of features dropped by stability selection; $J_\mathrm{R}$ is the Jaccard stability of RobustModelMaker's selected sets across outer folds. Paired $p$ is from the Wilcoxon signed-rank test on per-fold scores. Classification scores are AUC (higher is better); the regression score is RMSE (lower is better).}
\label{tab:cross-scenario}
\begin{tabular}{lcccccccc}
\hline
Scenario & Task & $n_\mathrm{train} \times d$ & $|\hat{S}|$ & Red. & $J_\mathrm{R}$ & Baseline & RobustModelMaker & $p$ \\
\hline
SECOM Manufacturing  & binary       & $1253 \times 590$ & 308 & 47.8\% & $0.660$ & $0.6251$ & $0.6548$ & $0.062$ \\
Urban Land Cover     & multiclass   & $540 \times 147$  &  56 & 61.9\% & $0.801$ & $0.9817$ & $0.9828$ & $0.125$ \\
Graphene Oxide Bulk  & regression   & $1293 \times 309$ & 138 & 55.3\% & $0.899$ & $0.0391$ & $0.0408$ & $0.625$ \\
\hline
\end{tabular}
\end{table}

\begin{table}[h]
\centering\small
\caption{Multi-method comparison across the three benchmark datasets. Each cell reports score $\pm$ standard deviation and the Jaccard stability $J$ of the selected feature sets across the five outer folds. Score conventions: AUC (binary), AUC-OVR (multiclass), RMSE in eV (regression). The baseline retains all features so its stability is undefined ($J = $ n/a). The retained-feature column reports $\bar{n}$, the mean number of selected features across folds, with the full feature count $d$ in brackets for reference.}
\label{tab:multi-method}
\small
\begin{tabular}{l c c c c}
\hline
Method & SECOM & Urban & Graphene & $\bar{n}$  \\
 & (AUC) &  (AUC-OVR) &  (RMSE, eV) & (SECOM / Urban / Graphene) \\
\hline
Baseline (all features)        & $0.6251 \pm 0.0364$ / n/a   & $0.9817 \pm 0.0037$ / n/a   & $0.0391 \pm 0.0215$ / n/a   & 590 / 147 / 309 \\
RobustModelMaker               & $0.6548 \pm 0.0477$ / $0.660$ & $0.9828 \pm 0.0034$ / $0.801$ & $0.0408 \pm 0.0227$ / $0.899$ & 308 / 56 / 138 \\
ANOVA $F$-test                 & $0.6659 \pm 0.0256$ / $0.417$ & $0.9726 \pm 0.0037$ / $1.000$ & $0.0202 \pm 0.0157$ / $1.000$ & 59 / 14 / 30 \\
RFECV                          & $0.6701 \pm 0.0572$ / $0.274$ & $0.9832 \pm 0.0045$ / $0.468$ & $0.0124 \pm 0.0154$ / $0.520$ & 254 / 85 / 3 \\
Boruta                         & $0.6188 \pm 0.0737$ / $0.224$ & $0.9825 \pm 0.0044$ / $0.895$ & $0.0347 \pm 0.0208$ / $0.958$ & 10 / 103 / 115 \\
\hline
\end{tabular}
\end{table}

Four observations follow from the cross-scenario pattern.

First, every method on every dataset is classified as \texttt{preserved} relative to the full-feature baseline. This is in part a structural consequence of the test: with $K = 5$ outer folds, the minimum two-sided $p$-value of the Wilcoxon signed-rank test is $1/16 \approx 0.0625$, so no paired comparison at this fold count can ever cross $\alpha = 0.05$ regardless of how decisive the per-fold ordering is. On SECOM, RobustModelMaker exceeds the baseline on every one of the five outer folds and is supported by a paired $t$-test $p = 0.019$, but the rank-based test returns $p = 0.0625$ and the outcome label remains \texttt{preserved}. On Graphene Oxide Bulk, ANOVA and Boruta each have all five fold scores better than the baseline ($p = 0.0625$) and would be classified \texttt{sig. better *} at any fold count of six or more; the present configuration cannot resolve this. On Urban Land Cover, no method is meaningfully separated from the baseline at any fold count. The headline finding is therefore that no selector tested here imposes a statistically detectable cost on predictive performance at the given fold structure, with the caveat that the test has a limited dynamic range at $n = 5$.

Second, on the multi-method comparison (Table~\ref{tab:multi-method}), RobustModelMaker is competitive in score on all three datasets but is not the highest-scoring selector on any of them. On SECOM and Urban Land Cover, the four selecting methods occupy a tight band of approximately 0.01--0.05 in AUC, well within fold-to-fold noise. On Graphene Oxide Bulk, ANOVA and RFECV achieve substantially lower RMSE than RobustModelMaker by exploiting the dataset's univariate structure, in which a small number of composition descriptors dominate the regression signal. This is a structural property of the dataset rather than a deficiency of the framework, and is reported as such.

Third, on the stability axis, RobustModelMaker is consistently in the upper half of the methods compared. On SECOM, where Boruta ($J = 0.224$) and RFECV ($J = 0.274$) are unstable to a degree that calls into question whether their selected sets would replicate, RobustModelMaker's $J = 0.660$ is the highest of any method that retains more than a fraction of the features. On Urban Land Cover, RobustModelMaker's $J = 0.801$ is second only to ANOVA's mechanically maximal $J = 1.000$ at $k = 14$. On Graphene Oxide Bulk, three methods (ANOVA, Boruta, RobustModelMaker) cluster at $J > 0.89$, with ANOVA mechanically perfect at $k = 30$. RFECV is the least stable selector on all three datasets, illustrating the failure mode that motivates bootstrap aggregation.

Fourth, the joint position on the score; stability frontier is the framework's actual claim. ANOVA achieves perfect stability by retaining a small fixed-size subset that may or may not contain the most informative features; on SECOM and Urban Land Cover its score is below RobustModelMaker's, while on Graphene Oxide Bulk its score is much better because the dataset rewards its univariate ranking. RFECV achieves competitive or leading score on all three datasets but at stability that ranges from low to unreliable. Boruta varies markedly between datasets and is brittle under severe imbalance. RobustModelMaker delivers a score that is within fold-to-fold noise of the best selector on each dataset, together with stability that is consistently in the top two across all three task types. The framework's value is the joint guarantee, not dominance on either axis alone.

The benchmark does not establish that RobustModelMaker is the best feature selector under all conditions. It establishes that, in the regime for which it is designed, small-to-medium scientific datasets where reproducibility and honest performance are both required outputs, it occupies a position on the score--stability frontier that none of the alternatives jointly match.

\section{Example Applications}
\label{sec:applications}

The benchmark suite establishes the framework's average-case behaviour under controlled, adversarial conditions. This section reports its use in two scientific applications where the analysis question is not benchmark-style comparison but a concrete predictive and interpretive task on real data. Both applications use the framework as it would be used in practice: a single configuration appropriate to the dataset, with the result evaluated on its own merits.

\subsection{Ovarian Cancer Biomarker Discovery}

\begin{figure}[h]
\centering
\begin{subfigure}[t]{0.3\linewidth}\centering
  \includegraphics[width=\linewidth]{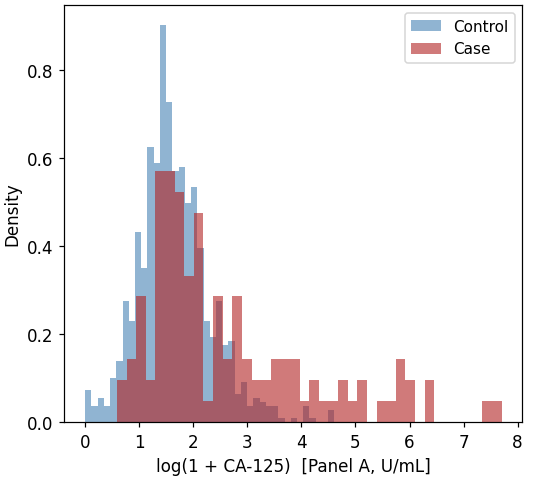}
  \caption{Class distribution}\label{fig:plco-a}
\end{subfigure}\hfill
\begin{subfigure}[t]{0.33\linewidth}\centering
  \includegraphics[width=\linewidth]{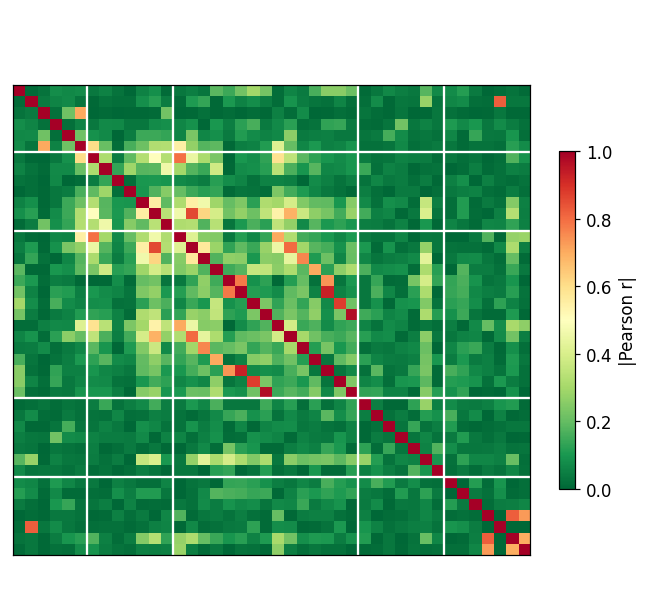}
  \caption{Feature-panel correlation}\label{fig:plco-b}
\end{subfigure}\hfill
\begin{subfigure}[t]{0.32\linewidth}\centering
  \includegraphics[width=\linewidth]{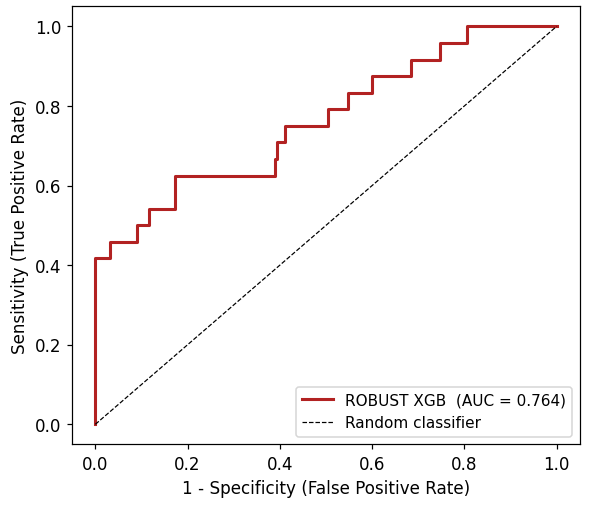}
  \caption{ROC on held-out test set}\label{fig:plco-c}
\end{subfigure}

\vspace{4pt}

\begin{subfigure}[t]{0.38\linewidth}\centering
  \includegraphics[width=\linewidth]{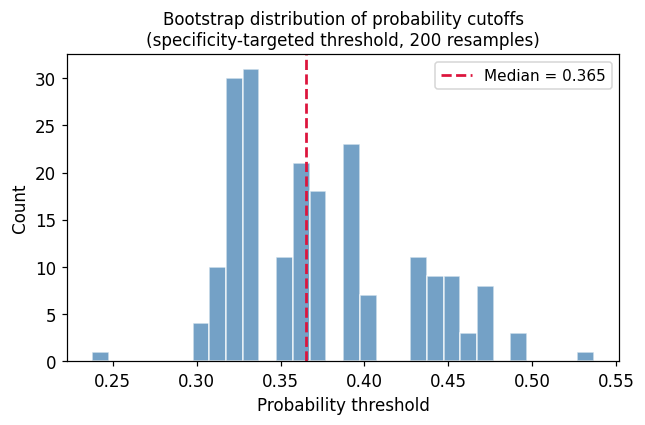}
  \caption{Bootstrap cutoff distribution}\label{fig:plco-d}
\end{subfigure}\hfill
\begin{subfigure}[t]{0.32\linewidth}\centering
  \includegraphics[width=\linewidth]{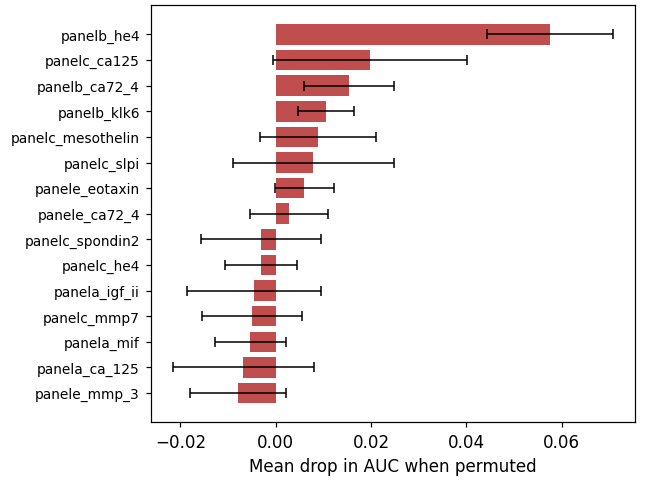}
  \caption{Permutation importance (top 15)}\label{fig:plco-e}
\end{subfigure}\hfill
\begin{subfigure}[t]{0.28\linewidth}\centering
  \includegraphics[width=\linewidth]{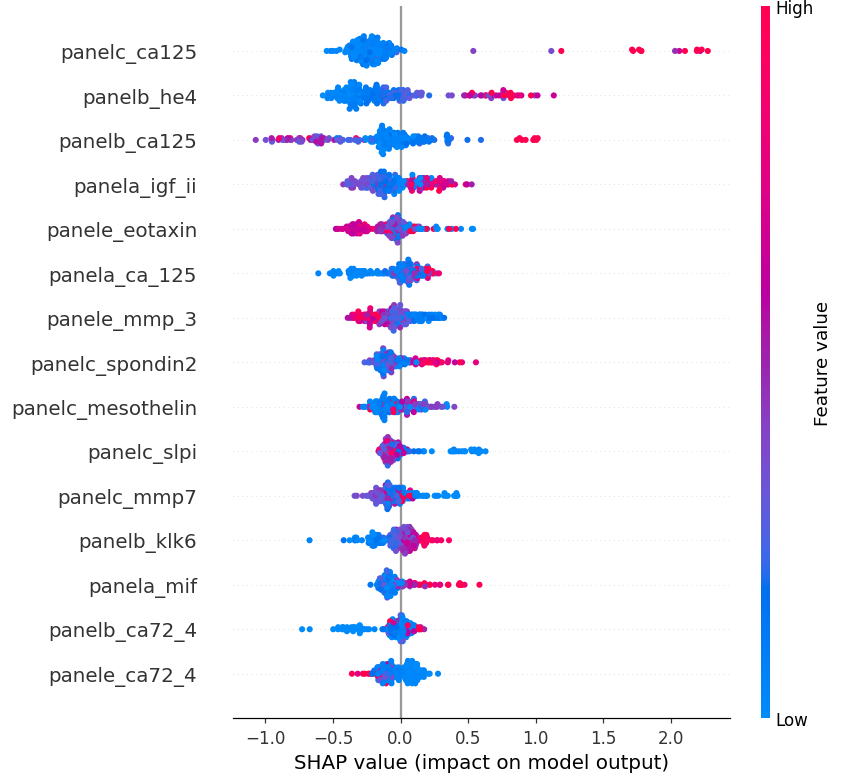}
  \caption{SHAP summary}\label{fig:plco-f}
\end{subfigure}
\caption{PLCO ovarian cancer biomarker discovery, all standard outputs of the analysis notebook. (a)~The cohort is heavily class-imbalanced: 118 cases against 951 matched controls (positive rate $\approx 11\%$). (b)~Feature-panel correlation heatmap showing the strong within-panel and cross-panel redundancy structure that motivates stability selection: the same biomarker is measured under different assay platforms across multiple panels (white lines). (c)~ROC curve on the held-out test set, AUC $= 0.764$, slightly above the nested cross-validation estimate of $0.737 \pm 0.027$; the agreement confirms that the nested CV estimate was not optimistically biased. (d)~Bootstrap distribution of probability cutoffs at the targeted specificity of $0.98$, from 200 resamples of the control-class out-of-fold predicted probabilities; median cutoff $0.365$, bulk of the distribution in $[0.30, 0.48]$. The spread is the explicit uncertainty in the operating point. (e)~Permutation importance on the held-out test set for the 16 selected biomarkers, ranked by mean drop in AUC under feature permutation; \texttt{panelb\_he4}, \texttt{panelc\_ca125}, and \texttt{panelb\_ca72\_4} are the leading contributors after selection. (f)~SHAP summary plot for the same selected feature set, complementing the permutation ranking with per-sample contribution sign and magnitude.}
\label{fig:plco}
\end{figure}

The first application is biomarker discovery on serum samples from the Prostate, Lung, Colorectal and Ovarian Cancer Screening Trial (PLCO)~\cite{prorok2000plco}. The dataset comprises 1,069 serum samples (118 ovarian cancer cases and 951 matched controls) with 42 biomarker concentrations measured across five immunoassay panels. The same biomarker often appears in more than one panel under different assay platforms (CA-125 appears in panels A, B, C, and E; HE4 appears in panels B, C, and E), and panel C also contains log-transformed variants of its raw measurements. The feature space is therefore characterised by substantial within-panel and cross-panel redundancy, which is precisely the setting where stability selection is informative because it identifies which specific assay version of a given biomarker is consistently discriminative. Figure~\ref{fig:plco}(a) shows the class distribution and Figure~\ref{fig:plco}(b) shows the absolute pairwise correlation across the 42 biomarker columns; the dense off-diagonal structure visualises the redundancy that motivates the use of stability selection on this dataset.

RobustModelMaker was run with XGBoost as the base algorithm, $n_\mathrm{bootstrap} = 100$ bootstrap iterations, stability threshold $\tau = 0.60$, $K = L = 5$ outer and inner cross-validation folds, $n_\mathrm{iter} = 30$ hyperparameter configurations per inner fold, and a cutoff bootstrap count of 200 with a target specificity of 0.98 for the binary cutoff. The choice of XGBoost over linear alternatives reflects the non-linear interactions among biomarker concentrations that are characteristic of the underlying biology. The configuration was chosen for an analysis exploring the feature subset rather than for publication-grade rigour; the framework's guidance suggests increasing $n_\mathrm{bootstrap}$ to 200 and $K = L = 10$ for publication-quality runs.

The framework selected 16 of the 42 biomarkers (62\% reduction). The nested cross-validation ROC-AUC on the training partition was $0.7370 \pm 0.0267$, with per-fold AUCs of $0.738$, $0.736$, $0.785$, $0.721$, and $0.705$. The bootstrap-determined probability cutoff at the target specificity of $0.98$ was $0.365$, with a 95\% confidence interval of $[0.308, 0.475]$; the achieved specificity at the cutoff was $97.9\%$ and the achieved sensitivity was $30.9\%$, both computed on the out-of-fold predictions. Sensitivity at this stringent specificity is the expected operating point for an early-detection screening test, where false positives carry substantial cost. Figure~\ref{fig:plco}(c)--(d) show the corresponding ROC curve on the held-out test set and the bootstrap distribution of the targeted-specificity cutoff. Figure~\ref{fig:plco}(e)--(f) show post-hoc per-feature contributions to the fitted model: permutation importance ranks features by mean drop in AUC under feature permutation on the held-out set, while the SHAP summary attributes per-sample contributions with sign and magnitude. The two views are consistent in their identification of \texttt{panelb\_he4}, \texttt{panelc\_ca125}, and \texttt{panelb\_ca72\_4} as the largest contributors to the fitted model, supporting the interpretation that the stability-selected panel is anchored on biomarkers with known clinical relevance to ovarian cancer.

The selected biomarker panel, summarised in Table~\ref{tab:plco}, recovers CA-125 in three different panel versions (panelA, panelB, panelC) and HE4 in two (panelB, panelC), demonstrating that the stability selection does not collapse correlated assay measurements into a single representative but retains the versions that are independently discriminative across bootstrap resamples. Mesothelin, MMP7, spondin-2, and SLPI from panel C, and eotaxin and MMP-3 from panel E, also appear in the selected set. The result is a 16-biomarker panel that is interpretable in clinical terms and assayable across the five panels without requiring all 42 measurements.

\begin{table}[h]
\centering
\caption{Selected biomarker subset on the PLCO ovarian cancer dataset (16 of 42 biomarkers, 62\% reduction). Biomarkers are grouped by the assay panel from which they were drawn.}
\label{tab:plco}
\begin{tabular}{ll}
\toprule
Panel & Selected biomarkers \\
\midrule
A (growth factors / cytokines)        & IGF-II, MIF, CA-125 \\
B (cancer antigens + HE4)             & CA-125, CA72-4, KLK6, HE4 \\
C (mesothelin panel)                  & CA-125, HE4, mesothelin, MMP7, spondin-2, SLPI \\
E (cytokines / CA-125 retest)         & eotaxin, MMP-3, CA72-4 \\
\bottomrule
\end{tabular}
\end{table}

\subsection{Superconductor Critical Temperature}

\begin{figure}[h]
\centering
\begin{subfigure}[t]{0.35\linewidth}\centering
  \includegraphics[width=\linewidth]{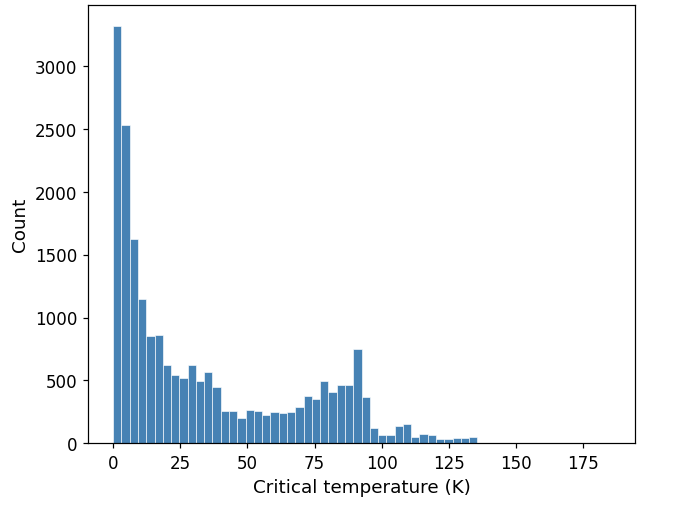}
  \caption{Target distribution}\label{fig:supercon-a}
\end{subfigure}\hfill
\begin{subfigure}[t]{0.32\linewidth}\centering
  \includegraphics[width=\linewidth]{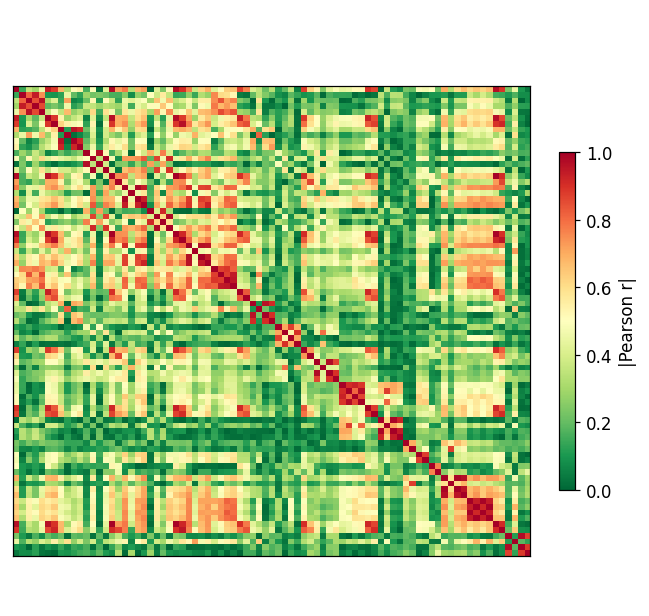}
  \caption{Feature correlation heatmap}\label{fig:supercon-b}
\end{subfigure}\hfill
\begin{subfigure}[t]{0.28\linewidth}\centering
  \includegraphics[width=\linewidth]{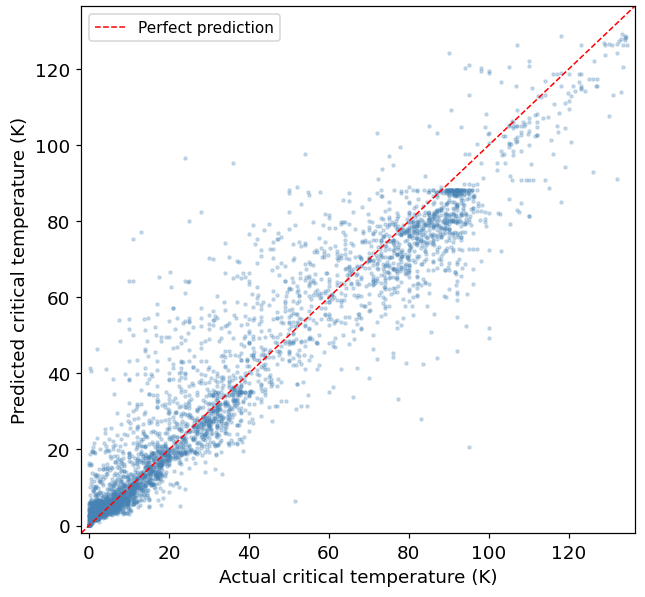}
  \caption{Predicted vs actual on test set}\label{fig:supercon-c}
\end{subfigure}

\vspace{4pt}

\begin{subfigure}[t]{0.28\linewidth}\centering
  \includegraphics[width=\linewidth]{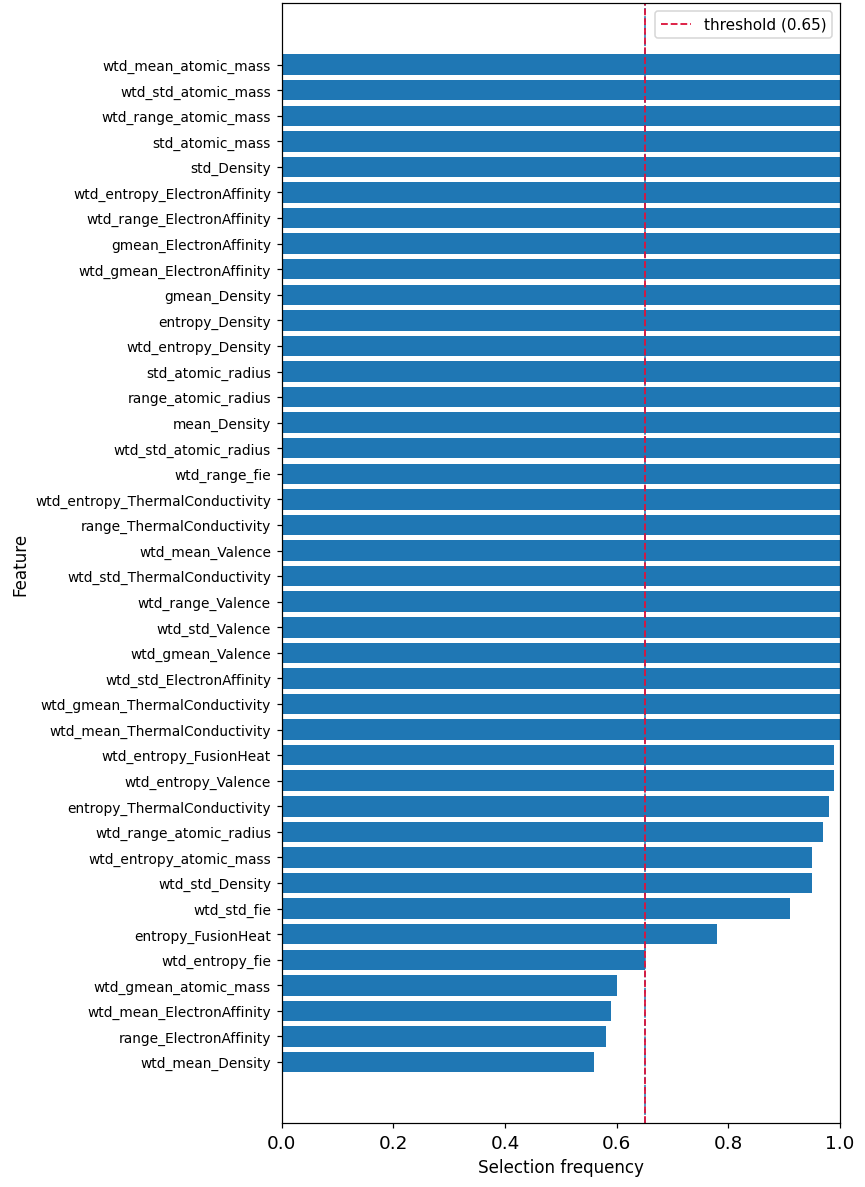}
  \caption{Bootstrap stability frequencies}\label{fig:supercon-d}
\end{subfigure}\hfill
\begin{subfigure}[t]{0.37\linewidth}\centering
  \includegraphics[width=\linewidth]{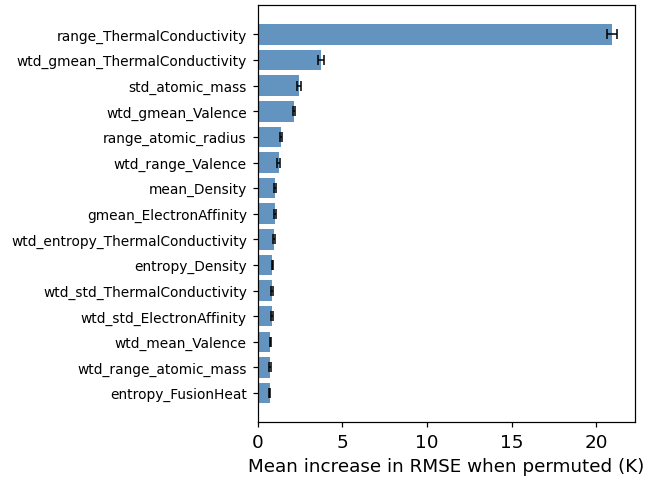}
  \caption{Permutation importance (top 15)}\label{fig:supercon-e}
\end{subfigure}\hfill
\begin{subfigure}[t]{0.34\linewidth}\centering
  \includegraphics[width=\linewidth]{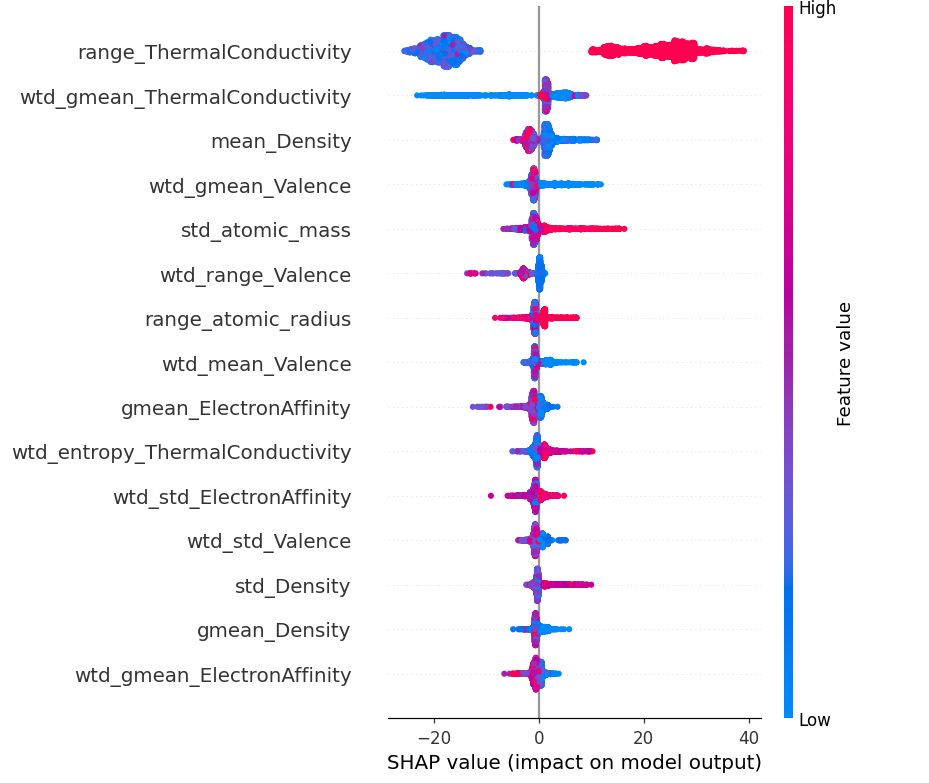}
  \caption{SHAP summary}\label{fig:supercon-f}
\end{subfigure}
\caption{Superconductor critical-temperature regression, all standard outputs of the analysis notebook. (a)~Distribution of the regression target (critical temperature in Kelvin) across the 21,263 superconductors; the distribution is heavily right-skewed with a mode near zero and a long upper tail past 100~K. (b)~Absolute pairwise correlation heatmap for the 81 composition-derived features, showing the dense block structure that arises from computing ten statistics per atomic property; ten-feature blocks along the diagonal correspond to one underlying physical quantity each. (c)~Predicted versus actual critical temperature on the held-out test set; the diagonal corresponds to perfect prediction, RMSE on the held-out test set is $10.1$~K and $R^2 = 0.912$, in agreement with the nested cross-validation estimate of $10.78 \pm 0.16$~K. (d)~Bootstrap selection frequencies $\pi_k$ for the top 40 features, output of the framework's \texttt{plot\_feature\_stability} method on the supercon notebook result. The dashed line marks the stability threshold $\tau = 0.65$. A dense cluster of features is selected in nearly every bootstrap run, a transition zone sits close to the threshold, and a tail of features falls below it. The 36 features retained in $\hat{S}$ are those to the right of the threshold line. (e)~Permutation importance on the held-out test set for the 36 selected features, ranked by mean increase in RMSE under feature permutation; \texttt{range\_ThermalConductivity} dominates, followed by \texttt{wtd\_gmean\_ThermalConductivity} and \texttt{std\_atomic\_mass}. (f)~SHAP summary plot for the same selected feature set, identifying both the most influential features and the sign of their per-sample contributions.}
\label{fig:supercon}
\end{figure}

The second application is regression of the critical temperature of superconducting materials from their composition statistics. The dataset is the UCI Superconductivity Data~\cite{hamidieh2018superconductor}, comprising 21,263 superconductors described by 81 composition-derived features. Each feature is one of ten statistics (mean, weighted mean, geometric mean, weighted geometric mean, entropy, weighted entropy, range, weighted range, standard deviation, weighted standard deviation) of one of eight underlying atomic properties (atomic mass, first ionisation energy, atomic radius, density, electron affinity, fusion heat, thermal conductivity, valence), plus the number of elements in the formula. The feature space is therefore strongly correlated within each property group by construction; the question of interest is which statistics of which properties carry independent predictive signal. Figure~\ref{fig:supercon}(a) shows the marginal distribution of the regression target, which is heavily right-skewed with a mode near zero and a long upper tail past 100~K. Figure~\ref{fig:supercon}(b) is the absolute pairwise correlation heatmap for the 81 features; the eight ten-feature blocks along the diagonal correspond to the eight atomic properties from which the statistics are derived, and the strong within-block correlation is the structural pattern that stability selection is asked to resolve.

The training and test data were partitioned by stratified 80/20 random split (17,010 train, 4,253 held-out test). RobustModelMaker was run with RF as the base algorithm, $n_\mathrm{bootstrap} = 100$, stability threshold $\tau = 0.65$, $K = L = 5$, and $n_\mathrm{iter} = 50$ hyperparameter configurations per inner fold. The framework selected 36 of the 81 features (55.6\% reduction). The nested cross-validation RMSE was $10.78 \pm 0.16$ K, with per-fold RMSEs of $10.97$, $10.89$, $10.53$, $10.69$, and $10.84$ K. The held-out test RMSE was $10.06$ K, with $R^2 = 0.912$ and MAE $= 6.22$ K. The agreement between the nested cross-validation estimate and the held-out test score confirms that the nested cross-validation estimate was honest. Figure~\ref{fig:supercon}(c) plots predicted against actual critical temperature on the held-out test set, and Figure~\ref{fig:supercon}(d) shows the bootstrap selection frequencies of the top 40 features with the stability threshold $\tau = 0.65$ marked; the per-fold concordance reported above is the empirical version of the diagonal pattern visible in the scatter.

A matched full-feature nested cross-validation baseline using the same algorithm, the same preprocessing, and the same five outer folds achieved RMSE $= 9.71 \pm 0.24$ K on 81 features. RobustModelMaker has a higher mean RMSE by $1.07$ K, with the baseline scoring lower on all five outer folds. This is the explicit trade-off the framework is designed to make visible: the stability-selected feature set is more reproducible and substantially smaller, but the baseline using all 81 features produces marginally lower error on this dataset. Whether the trade is worth making depends on the downstream use case. For a model that will be deployed and re-fitted on similar data in the future, the smaller and stable subset is preferable; for a one-off prediction task on this exact dataset, the baseline is preferable.

Figure~\ref{fig:supercon}(c) shows the full distribution of bootstrap selection frequencies for all 81 features, with the threshold marked. The distribution is bimodal in the relevant range: a substantial cluster of features is selected in nearly every bootstrap run, a smaller group falls just above the threshold, and a long tail of features lies below it. This bimodality is the structural signature that stability selection is designed to expose: it identifies the features that survive bootstrap resampling cleanly and separates them from those whose selection is contingent on the particular training sample.

Table~\ref{tab:supercon} groups the 36 selected features by their underlying atomic property and reports the count of selected statistics per property. The distribution is informative: density, thermal conductivity, and electron affinity contribute the largest number of stable statistics (six, six, and five respectively), atomic mass, atomic radius, and valence contribute moderate numbers, and fusion heat contributes only two. The number of elements in the formula was not selected. Across all eight properties, the weighted statistics (those that incorporate the elemental stoichiometry of the compound) are systematically more frequently selected than the unweighted versions, which is consistent with the physics of property-composition relationships in mixed materials. Figure~\ref{fig:supercon}(e) and (f) are the post-hoc per-feature contribution panels: permutation importance on the held-out test set and the corresponding SHAP summary, both operating on the 36 stability-selected features. Both views identify \texttt{range\_ThermalConductivity} as the single largest contributor by a wide margin, followed by \texttt{wtd\_gmean\_ThermalConductivity} and \texttt{std\_atomic\_mass}; the agreement between the two post-hoc methods reinforces the interpretation that thermal conductivity statistics dominate the regression signal in this composition representation.

\begin{table}[h]
\centering
\caption{Selected feature counts on the UCI Superconductivity dataset, grouped by the underlying atomic property. RobustModelMaker retained 36 of 81 features.}
\label{tab:supercon}
\begin{tabular}{lcc}
\toprule
Atomic property & Selected statistics & Total available \\
\midrule
Atomic mass            & 5 & 10 \\
First ionisation energy & 3 & 10 \\
Atomic radius          & 4 & 10 \\
Density                & 6 & 10 \\
Electron affinity      & 5 & 10 \\
Fusion heat            & 2 & 10 \\
Thermal conductivity   & 6 & 10 \\
Valence                & 5 & 10 \\
Number of elements     & 0 & 1  \\
\midrule
Total                  & 36 & 81 \\
\bottomrule
\end{tabular}
\end{table}

A direct comparison with the published results of Nawoda et al.~\cite{nawoda2026superconductor} on the same target task provides useful context. Nawoda et al. assembled a balanced dataset of 13,415 superconductors and 13,425 non-superconductors (26,840 total) from the SuperCon database augmented with six non-superconductor databases. They engineered 212 features, including 198 statistical descriptors, CuO-layer-specific weights, per-element thermal conductivities, and a material-type label distinguishing low-temperature superconductors, high-temperature superconductors, and non-superconductors. Embedded random forest importance with a cumulative-importance threshold of 98\% retained 72 of the 212 features. They evaluated 14 algorithms under a single random 80/20 split, finding their best random forest result at MAE $= 1.748$ K, $R^2 = 0.962$ without tuning (with the authors noting overfitting at Train MAE $= 0.656$) and MAE $= 3.386$ K, $R^2 = 0.919$ with depth-6, 200-tree tuning. Their best overall model was LightGBM at MAE $= 2.933$ K, $R^2 = 0.930$.

Direct numerical comparison is complicated by three structural differences. The Nawoda et al. dataset is approximately a quarter larger than the UCI dataset and includes non-superconductors with $T_c = 0$ K which are easy to predict and deflate aggregate error metrics. Their single most important feature by random forest importance was the material-type label, which contributes approximately 60\% of total importance and is not available in the UCI dataset. Their evaluation protocol is a single random 80/20 split rather than nested cross-validation. Controlling for these factors, the analysis presented here achieves $R^2 = 0.912$ on a strictly held-out test set with a leakage-safe nested cross-validation estimate that the test result confirms, while additionally providing a stability-tested feature subset and an explicit account of feature reproducibility across bootstrap resamples. The single-split protocol used by Nawoda et al. cannot provide either.

\subsection{Discussion}

The two applications illustrate the framework operating in two regimes that differ in dataset size, task type, and the relationship between selected features and scientific interpretation. The PLCO case is the regime in which the framework's design priorities are most directly load-bearing: a modest sample size (1,069), strong within-panel feature redundancy, a clinically meaningful cutoff requirement, and an analysis output (a biomarker panel) whose downstream use depends on its reproducibility. The framework produces a 16-biomarker panel and a probability cutoff with an explicit bootstrap confidence interval, and reports its sensitivity and specificity at that cutoff without further adjustment. The 30.9\% sensitivity at 98\% specificity is not an aspirational number; it is the operating point that the data support at the chosen specificity target, and is reported as such.

The superconductivity case is the regime in which the framework's trade-off is most clearly visible. The full-feature baseline obtains lower RMSE than the stability-selected subset by approximately 1 K, on every fold, and the framework reports this delta rather than absorbing it into a wider variance interval. The case for the stability-selected subset is not that it predicts better, but that the 36 features it retains are robust under bootstrap resampling and the held-out test RMSE confirms the nested cross-validation estimate. The published comparison to Nawoda et al. clarifies what stability selection adds beyond a single-split analysis with embedded importance: a feature set that is itself a deliverable of the cross-validated procedure rather than a by-product of a single fit.

Across both cases, the framework's primary contribution is not raw predictive performance but the coupling of a reproducible feature subset with an honest performance estimate, reported with their associated uncertainties. The empirical results support the framework's design choices in their intended regime and make the trade-offs of those choices explicit where they appear.

\section{Conclusion}
\label{sec:conclusion}

RobustModelMaker is a framework for the regime in which two methodological failure modes of conventional machine learning pipelines, unstable single-run feature selection and the optimistic bias of non-nested cross-validation, jointly determine whether the analysis is replicable. The framework couples bootstrap stability selection with nested cross-validation in a fixed structural ordering, performs all preprocessing inside each outer fold, and produces a feature subset and a performance estimate that are derived from the same disjoint partition of the data. The principal claims of this paper are that the coupling is necessary rather than optional in this regime, that it can be implemented in a way that is both reproducible and computationally tractable for small-to-medium scientific datasets, and that the trade-offs it imposes on predictive performance are modest in practice and are reported transparently in the framework's outputs. The benchmark suite quantifies these trade-offs against three alternative selectors and shows that the framework's distinguishing property is its joint position on the score--stability frontier, not dominance on either axis alone.

The framework is not intended for streaming or out-of-core data, for very large datasets where throughput is the binding constraint, or for online learning. It is not a substitute for post-hoc interpretation tools such as SHAP, which answer a different question and operate on already-fitted models. It does not perform automated algorithm selection or ensembling across base learners. These restrictions are deliberate design choices, not omissions, and define the boundary of the regime in which the framework is the appropriate tool.

The framework is open-source, distributed via PyPI, and supported by a deterministic test suite spanning unit, performance, and reproducibility tests. The benchmark suite is included with the library and is reproducible by any user with the same environment. Two real-world applications, ovarian cancer biomarker discovery on the PLCO Trial dataset and superconductor critical-temperature regression on the UCI Superconductivity dataset, illustrate how the framework is used in practice and what trade-offs become visible when stability is treated as a first-class output rather than an emergent property of the model fit.

\section*{Availability}

RobustModelMaker is open-source and available at: \url{https://github.com/amaxiom/RobustModelMaker}, or via \texttt{pip install robustmodelmaker}.

\bibliographystyle{unsrt}
\bibliography{references.bib}

\end{document}